\documentclass[10pt,twocolumn,letterpaper]{article}

\usepackage{iccv}              
\usepackage[accsupp]{axessibility}
\usepackage{makecell}

\definecolor{iccvblue}{rgb}{0.21,0.49,0.74}
\usepackage[pagebackref,breaklinks,colorlinks,allcolors=iccvblue]{hyperref}

\usepackage{algorithm}
\usepackage{algpseudocode}
\usepackage{amssymb}
\usepackage{xcolor, pifont}
\newcommand{\mydowntriangle}{\textcolor{blue}{\ding{116}}}
\newcommand{\mycircle}{\textcolor{green}{\ding{108}}}
\newcommand{\myx}{\textcolor{red}{\ding{54}}}
\usepackage{tabularx}
\usepackage{xr}

\def\paperID{13947}
\def\confName{ICCV}
\def\confYear{2025}

\title{Granular Concept Circuits: \\ Toward a Fine-Grained Circuit Discovery for Concept Representations
}

\usepackage[symbol]{footmisc}

\author{
Dahee Kwon\footnotemark[1] \\
KAIST AI \\
{\tt\small daheekwon@kaist.ac.kr}
\and
Sehyun Lee\footnotemark[1] \\
KAIST AI \\
{\tt\small sehyun.lee@kaist.ac.kr}
\and
Jaesik Choi\footnotemark[2] \\
KAIST AI, INEEJI \\
{\tt\small jaesik.choi@kaist.ac.kr}
}

\begin{document}

\makeatletter
\let\@oldmaketitle\@maketitle
\renewcommand{\@maketitle}{\@oldmaketitle
\centering
  \includegraphics[width=0.93\textwidth]
    {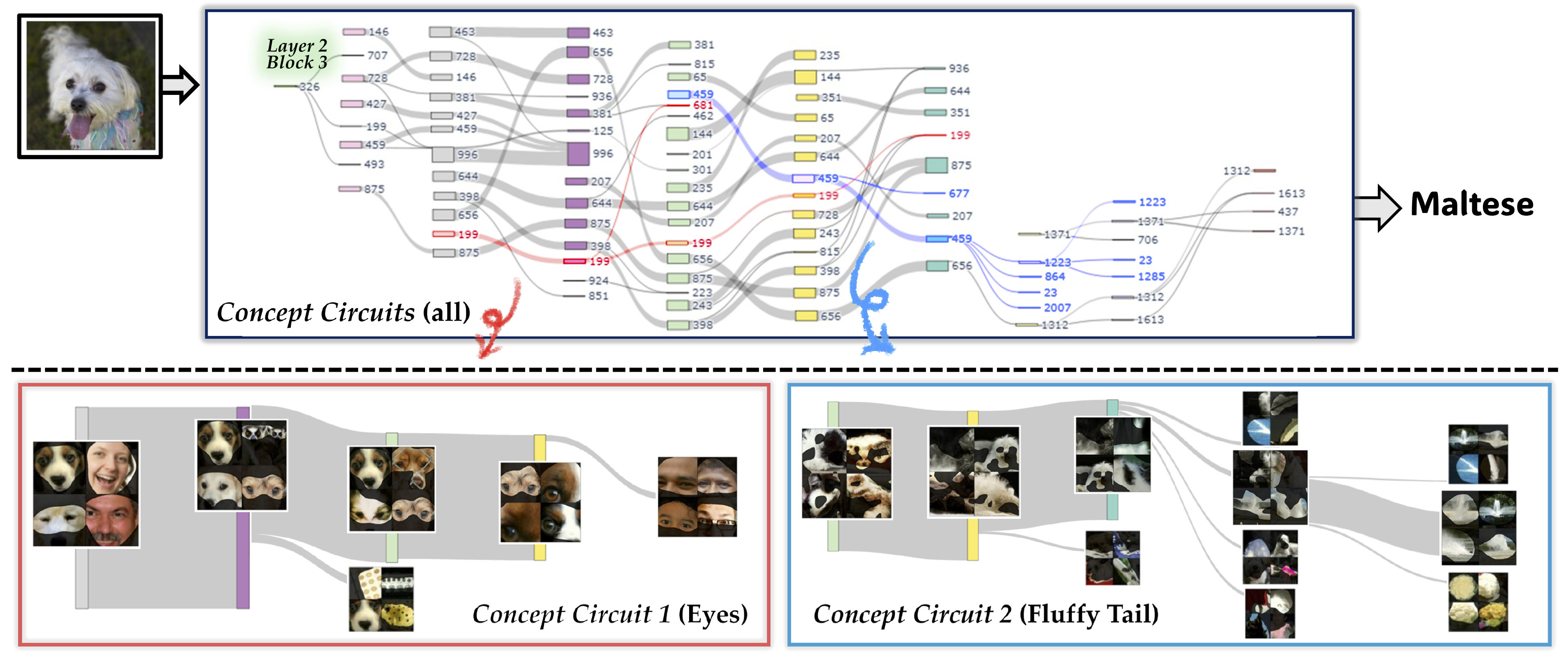}
     \captionof{figure}{\textbf{Granular Concept Circuits}. \textbf{(Top)} Circuits obtained by merging all captured 20 circuits conditioned on the query image in ResNet50. \textbf{(Bottom)} Examples of individual concept circuits. Edge thickness represents the connectivity between neurons computed with the proposed Neuron Sensitivity Score. More examples are depicted in~\cref{fig:multi-concepts}.} 
         \label{fig:page_one}
         \bigskip}
\makeatother
\maketitle

\footnotetext[1]{Equal contribution.}
\footnotetext[2]{Corresponding author.}

\begin{abstract}
Deep vision models have achieved remarkable classification performance by leveraging a hierarchical architecture in which human-interpretable concepts emerge through the composition of individual neurons across layers. Given the distributed nature of representations, pinpointing where specific visual concepts are encoded within a model remains a crucial yet challenging task. In this paper, we introduce an effective circuit discovery method, called \textit{Granular Concept Circuit (GCC)}\footnote{Code is available at \url{https://github.com/daheekwon/GCC}.}, in which each circuit represents a concept relevant to a given query. To construct each circuit, our method iteratively assesses inter-neuron connectivity, focusing on both functional dependencies and semantic alignment. By automatically discovering multiple circuits, each capturing specific concepts within that query, our approach offers a profound, concept-wise interpretation of models and is the first to identify circuits tied to specific visual concepts at a fine-grained level. We validate the versatility and effectiveness of GCCs across various deep image classification models.

\end{abstract}    

\section{Introduction}
\label{sec:intro}
\vspace{-0.2em}
In deep image classification models, representations evolve hierarchically—from low-level features like points and edges in early layers, to mid-level patterns such as contours and textures, and finally to high-level semantic representations of objects and scenes~\cite{bau2017network, bau2020units, olah2020overview, dorszewski2025colors}. Throughout this hierarchy, concepts, defined as meaningful representations comprehensible to humans, emerge through the composition of individual neurons, with each neuron contributing to progressively abstract representations~\cite{hinton1984distributed, hinton1986learning}. 

However, much of the existing research in deep learning interpretation has focused on associating individual neurons with specific concepts~\cite{bau2017network, bau2020units, kalibhat2023identifying, hernandez2021natural}. These studies typically interpret neurons in isolation without explicitly considering relational structures among them. While this single-neuron approach provides valuable insights into isolated concept representations, it inherently overlooks the distributed nature of neural encoding, where concepts are inherently represented across multiple neurons and layers. Indeed, this form of distributed representation closely resembles how the human brain stores and recalls information~\cite{rumelhart1986parallel, wood1996neuroscience, hebb2005organization}, leading to extensive studies on the mechanisms underlying these neural connections~\cite{bennett2018rewiring,jeong2021synaptic,schmidt2022genetic,rissman2012distributed}. Building upon these neuroscience-inspired insights and the previously discussed neuron interpretability methods, recent research has begun to trace how visual concepts are represented across multiple neurons and layers.

For example, VCC~\cite{kowal2024visual} represents layer-wise concepts using Concept Activation Vectors (CAVs) and analyzes their connectivity; however, its misalignment with the network structure makes it difficult to localize where specific concepts emerge precisely. \citet{rajaram2024automatic}, hereafter referred to as ADVC, extend the connectivity-based methods of \citet{olah2020overview} by iteratively computing and refining the connectivity strength between adjacent layers. Nevertheless, these methods remain constrained by their focus on class-relevant relationships, limiting deeper understanding and interpretation of the models. To analyze models more comprehensively, it is essential to capture concepts beyond class-specific ones with finer granularity and identify where and how each concept is encoded. Importantly, existing approaches do not aim to subdivide these concept representations or analyze them in a fine-grained manner, fundamentally differing from our objective.

\begin{table}[!t]
\centering
\begin{tabular}{c|c|c|c|c}
\toprule \toprule
                        & CRP & VCC & ADVC & \textbf{GCC} \\ \midrule
Circuit               & \myx    & \mycircle & \mycircle & \mycircle      \\ \midrule
Neuron-based    & \mycircle    & \myx & \mycircle & \mycircle  \\ \midrule
Concept Specificity & -  & \myx  & \myx & \mycircle \\ \midrule
$\#$ of Input       & 1   & class   & few & versatile \\ \midrule 
$\#$ of Output Circuit  & -  & 1    & 1 & $\geq$ 1     \\
\bottomrule \bottomrule

\end{tabular}
\caption{Comparative Analysis of CRP~\cite{achtibat2023attribution}, ADVC~\cite{rajaram2024automatic}, VCC~\cite{kowal2024visual}, and GCC (Ours).}
\label{tab:comparison}
\end{table}

Building on this, we introduce Granular Concept Circuits (GCC), an effective method for discovering visual concept circuits that leverage the distributed nature of representations to capture concepts relevant to a given query. To form a concept circuit, we assess inter-neuron connectivity by examining both their dependencies and semantic alignment, quantified using two scores: Neuron Sensitivity Score ($S_{NS}$) and Semantic Flow Score ($S_{SF}$). By iteratively identifying strongly connected neurons that represent coherent concepts, our method automatically constructs a visual concept circuit for each query. This results in multiple circuits, each encoding a distinct concept relevant to the query. The proposed circuit discovery method can be applied to various scenarios: from a single query to multiple queries, and from exploring common concepts to identifying unique concepts among them. For convenience, we refer to both the proposed method and the resulting circuit as Granular Concept Circuit (GCC). Our main contributions are summarized as follows:

\begin{itemize}
    \item We propose a method to automatically discover Granular Concept Circuits, where each circuit encodes a distinct concept relevant to the given query.
    \item We iteratively identify connected neurons based on two proposed scores, $S_{NS}$ and $S_{SF}$, to construct circuits representing coherent concepts.
    \item Our method enables flexible circuit construction based on the number of queries, concept types, and user-defined objectives.
\end{itemize}
\section{Related work}
\label{sec:related_work}

\begin{figure*}[!ht]
  \centering
     \includegraphics[width=\textwidth]{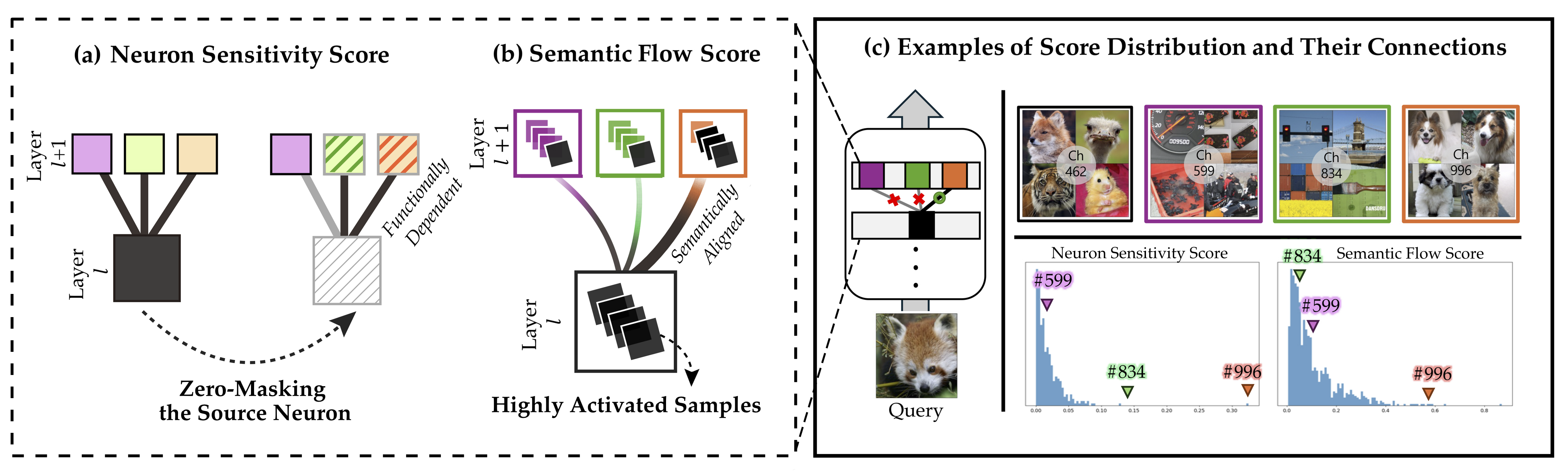}
     \caption{Illustration of Neuron Sensitivity Score ($S_{NS}$) and Semantic Flow Score ($S_{SF}$).  
    \textbf{(a)} $S_{NS}$ is computed by masking the source neuron and evaluating its connections (Eq.~\eqref{eq:neuron_sentitivity}).  
    \textbf{(b)} $S_{SF}$ quantifies semantic alignment across layers, with stronger alignment (orange) indicating stable concept interpretability (Eq.~\eqref{eq:semantic_flow}).  
    \textbf{(c)} Score distribution and corresponding neuron connections, with highly activated samples shown for selected neurons.  
    \textbf{Legend:} (\textcolor{purple}{purple}) low $S_{NS}$, low $S_{SF}$; (\textcolor{green}{green}) high $S_{NS}$, low $S_{SF}$; (\textcolor{orange}{orange}) high $S_{NS}$, high $S_{SF}$.  
    }
     \label{fig:score} 
\end{figure*}

\subsection{Concept-based Interpretability}
Analyzing visual concepts learned by deep neural networks has become essential for understanding and interpreting their decision-making processes. Previous methods can be broadly categorized based on whether they analyze concepts individually or explicitly examine relationships between multiple concepts.

\textit{Single-concept approaches} can be divided by how concepts are represented: neuron-based and vector-based approaches. Neuron-based methods—such as NetDissect~\cite{bau2017network}, CLIP-Dissect~\cite{oikarinen2022clip}, RDR~\cite{chang2024understanding}, FALCON~\cite{kalibhat2023identifying}, and MILAN~\cite{hernandez2021natural}—aim to identify the semantic meanings associated with individual or small groups of neurons. These methods offer intuitive interpretations but are often insufficient for explaining complex model decisions that involve interactions among multiple concepts. In contrast, vector-based methods—such as Concept Activation Vectors (CAVs)~\cite{kim2018interpretability, ghorbani2019towards}—represent concepts as activation directions spanning multiple neurons, enabling more abstract and generalizable representations. However, they often depend on manually labeled concept examples and struggle with capturing fine-grained semantics.

To address this limitation, \textit{relation-based approaches} have emerged, explicitly capturing interactions between multiple concepts. For example, HINT~\cite{wang2022hint} identifies collaborative neurons within a single layer to represent specific concepts using CAV, but the detailed mechanisms underlying their interactions across layers to form hierarchical representations remain uncertain. CRAFT~\cite{fel2023craft} leverages non-negative matrix factorization to trace and decompose concepts across multiple layers, but because it starts from the final classifier layer, it is limited to extracting concepts directly used for classification. CRP~\cite{achtibat2023attribution} computes neuron relevance scores based on conditional dependencies between adjacent layers, but its focus on pairwise layer-wise relevance limits its ability to capture circuit dynamics.

\subsection{Visual Circuit Discovery}
As interest in understanding relationships between concepts grows, researchers are increasingly exploring ways to identify these relationships across multiple layers with connection weightings, forming as a visual circuit. VCC~\cite{kowal2024visual} visualizes inter-layer connectivity by progressively decomposing concept vectors from a single class into finer components while measuring their relationships. Though it reveals concept structures across layers, it does not fully address how specific model components map to the learned concepts. The approach of interpreting a model through its components, known as mechanistic interpretability, offers insights into decision-making processes, key model components for specific tasks, and the origins of undesired behaviors. While extensively studied in LLMs~\cite{wang2022interpretability,meng2022locating,conmy2023towards, marks2024sparse}, its application in the vision domain remains limited. Early approaches primarily relied on manual analysis of model weights or neuron activations to extract meaningful representations~\cite{olah2020overview}, but scalability challenges make them impractical for large-scale models.

There has been an effort to discover visual circuits in partially automated way, (ADVC)~\cite{rajaram2024automatic}, where a circuit is identified to represent visual concepts from a few given examples. This method iteratively refines connectivity using cross-layer attribution, computed by multiplying activation values with gradients to capture neuron-to-neuron interactions in the model’s final decision. Since it constructs a single visual circuit, the resulting circuit unifies common concepts across queries. While intuitive, a deeper understanding of the model would benefit from a more granular decomposition of circuits according to each learning concept and an extension beyond class-related concepts to capture a wider range of attributes within an image. In this regard, our method advances visual circuit discovery by identifying circuits at the level of granular concept units. The comparison with other methods is organized in Table~\ref{tab:comparison}.

\section{Method}
\label{sec:method}
Our objective in this paper is to identify \textit{Granular Concept Circuit}—a set of neurons across multiple layers within a model that encodes a concept relevant to a given query. Instead of analyzing individual neurons, we focus on circuits that better capture the distributed nature of deep representations. This circuit-level view facilitates the identification of concepts, which are generally encoded through interactions among multiple components rather than within single units. While our method naturally extends to multiple queries, we describe it in the context of a single query for clarity. The extension to multiple samples is covered in Section~\ref{sec:exp}.

\subsection{Neuron Connectivity Identification}
\label{subsec:association}
A well-constructed circuit must exhibit meaningful connections among its components and consistently represent a common piece of information~\cite{yu2024functional,wang2022interpretability}. Based on these properties, we aim to assess the degree of connectivity between neurons\footnote{In this paper, we define a neuron as the smallest unit encoding spatial information, corresponding to a channel in CNN-based models and a hidden dimension in Transformer models.} that constitute a circuit.  

Let $\mathcal{F}$ denote the deep learning model under analysis, which can be expressed in a layer-wise manner: $\mathcal{F} = f^L \dots \circ f^l \dots \circ f^0$. The activation at layer $l$ of given query $x_q$ is given by $f^{l:0}(x_q) = a^l$.

Here, given a source neuron $a^l_c$, $c$-th neuron of layer $l$, we seek to determine the corresponding neuron in the next layer that should be connected to form a coherent concept circuit. To establish meaningful connections, we propose a Neuron Sensitivity Score $S_{NS}$, which quantifies the influence of a source neuron on a target neuron built on an intervention-based approach by measuring the change in the target’s activation when the source neuron is muted—i.e., set to zero~\cite{zhou2018revisiting, xin2019part,dai2021knowledge}. This operation captures both direct and indirect dependencies, reflecting how much the target neuron relies on the source. A high $S_{NS}$ indicates that the target neuron is strongly dependent on the information encoded in the source neuron, which we interpret as a functionally meaningful connection. To ensure we focus only on positively correlated relationships, we clip any negative score values to zero. While evaluating all combinations of source neurons would provide a more comprehensive causal attribution, the exponential complexity $O(2^{|N|})$ renders it intractable. Hence, here, we focus on a first-order approximation by measuring each node’s effect.

\begin{align}
\tilde{S}_{NS,c} &= \max\big(0, f^{l+1}(a^l_c)-f^{l+1}(\hat{a}^l_c) \big) \\
  & S_{NS}= \frac{\tilde{S}_{NS}}{\sum \tilde{S}_{NS}}
  \label{eq:neuron_sentitivity}
\end{align}

\noindent where $\hat{a}^l_c$ denotes layer $l$ activations with zero-masked $c-th$ node. Despite a high $S_{NS}$, the non-linearity of $f$ may still lead to spurious connections between semantically unrelated nodes (see Figure~\ref{fig:score}). Such connections are not only uninterpretable but also risk misrepresenting the actual node relationships. To address this, we introduce an additional constraint, Semantic Flow Score $S_{SF}$, to ensure semantic alignment of the encoded information between neurons as well as its strong dependency. This score is motivated by the idea that a neuron's encoded information is reflected in the set of samples where it is highly activated. We extract the top-$k$ highly activated samples for both the source and target neurons, and compute their overlap to quantify shared activation patterns.

\begin{equation}
    S_{SF} = \frac{|\mathcal{S}_{\text{src}} \cap  \mathcal{S}_{\text{tgt}}|}{|\mathcal{S}_{\text{src}}|}
    \label{eq:semantic_flow}
\end{equation}

\noindent where $\mathcal{S}_{\text{src}}$ denotes the set of highly activated samples for the source node, and $\mathcal{S}_{\text{tgt}}$ represents the corresponding set for the target node.  

Thus, we consider a source and target neuron to be connected—forming a building block of a concept circuit—when two criteria are met: (1) the target neuron exhibits strong functional dependency on the source, as indicated by a Neuron Sensitivity Score ($S_{NS} > \tau_{NS}$), and (2) it sufficiently preserves the information flowing from the source, as measured by a Semantic Flow Score ($S_{SF} > \tau_{SF}$).

\begin{algorithm}[t]
\caption{Finding a Single Circuit}
\label{alg:circuit}
\begin{algorithmic}[1]
\Require Query $x_q$, root node $r= (l, c)$, model $\mathcal{F}$, thresholds $\tau_{NS}$, $
\tau_{SF}$
\Ensure A concept circuit $circuit$

\State $circuit \gets \emptyset$  \Comment{\textbf{Initialize empty circuit}}
\State $N_s \gets \{r\}$  \Comment{\textbf{Initialize start nodes}}
\While {$N_s \neq \emptyset$} 
    \State $n_s \gets N_s.pop()$    

    \Statex \Comment{\textbf{Compute scores for next-layer node $i$}}
    \State $N_t \gets \Big\{ (n_s.l + 1, i) ~\Big| S_{NS}(n_s, x_q, f)_i \geq \tau_{NS}$
    \Statex \hspace{3.7cm} $\land \; S_{SF}(n_s, f)_i \geq \tau_{SF} \Big\}$
    \Statex \Comment{\textbf{Update circuit and search frontier}}
    \State $circuit \gets circuit \cup \{(n_s, n_t) \mid n_t \in N_t\}$ 

    \State $N_s \gets N_s \cup N_t$  
\EndWhile
\State \Return $circuit$
\end{algorithmic}
\end{algorithm}

\begin{figure*}[t]
  \centering
     \includegraphics[width=\textwidth]{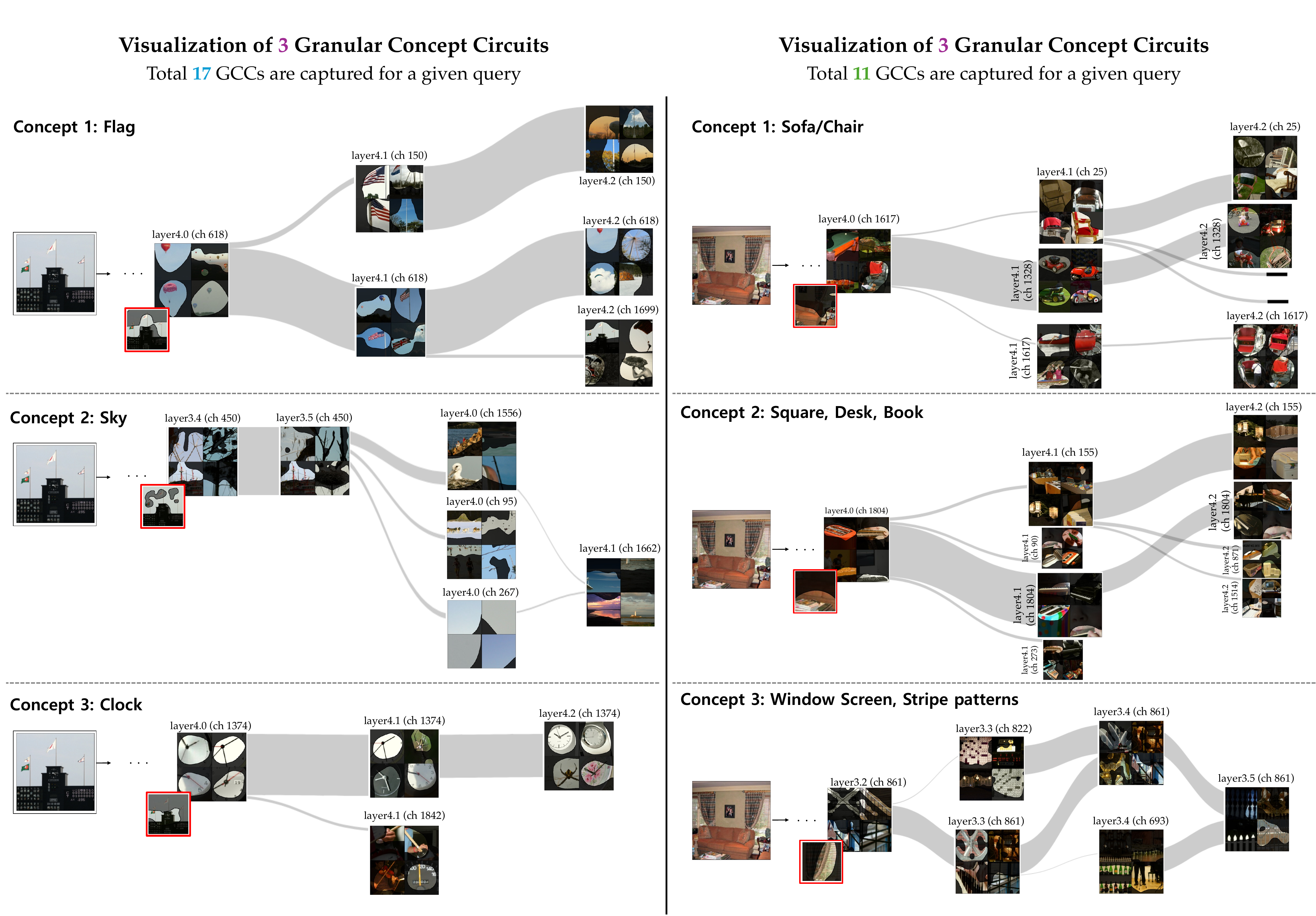}
     \caption{Qualitative results of \textbf{GCC} on ResNet50. For two given queries labeled as scoreboard and window screen, three distinct GCCs among all captured circuits are visualized for each query. Red solid outlines highlight the activated query regions corresponding to each circuit’s root node. }
     \label{fig:multi-concepts} 
\end{figure*}

\subsection{Granular Concept Circuit discovery}
By extending the connectivity described in~\cref{subsec:association} across multiple layers, we aim to uncover cross-layer connections that encode coherent concepts hierarchically. Here, a \textit{circuit} is a directed acyclic graph (DAG), where each node $n$ represents a neuron, and each edge $(n_s, n_t)$ denotes connectivity between adjacent layers. Discovering all concept circuits for a query involves four steps: (1) extracting root nodes $R = \{n_{r_1}, ..., n_{r_k}\}$ from a trained model $\mathcal{F}$ given a query $x_q$, (2) identifying connections from each root node to the next layer using the proposed scores, (3) iteratively tracking connectivity until no further connections exist, forming a Granular Concept Circuit, and (4) repeating this for all root nodes to construct the complete set of circuits.

The first step involves extracting a set of root nodes $R$. To begin, given a query $x_q$, we selectively explore only the nodes whose activations are sufficiently large comparable to other input samples. The idea that highly activated neurons carry significant information about the input query is well-supported by prior studies~\cite{bau2017network,achtibat2023attribution,kalibhat2023identifying}. In practice, we use nodes with activations' rank within the top 1$\%$ across all samples as root nodes.

In the second step, for each root node, we identify its connections to nodes in the next layer using the Neuron Sensitivity Score ($S_{NS}$) and Semantic Flow Score ($S_{SF}$), as described in~\cref{subsec:association}. After computing $S_{NS}$ for each candidate node in the adjacent layer, we extract those with sufficiently strong connectivity to the root node by applying a threshold $\tau_{NS}$. To automate this process and avoid manual threshold tuning, we apply the Peak-over-Threshold (POT) method~\cite{haan2006extreme, davison1990models, pickands1975statistical}, a statistical approach from extreme value theory. POT estimates an appropriate threshold by modeling the tail of the $S_{NS}$ distribution, providing a simple yet principled way to select highly responsive connections. See \Cref{sec:threshold} for details. Following this, we again filter out nodes that are less semantically aligned with root node, using the Semantic Flow Score $S_{SF}$. Our guideline for $\tau_{SF}$ is to use average score of all nodes, which guarantees the connected nodes at least has shared information with root note more than others on average. After filtering the neurons based on both conditions, the selected nodes are added to the circuit, with connections between neurons weighted by the $S_{NS}$ value.

Next, to explore further connections across layers, the newly added nodes serve as new starting points, and the process is iteratively repeated until no further connections remain. This results in a Granular Concept Circuit—a concept-specific circuit that originates from a root node. The full procedure is illustrated in~\cref{alg:circuit}.

Finally, in the last step, we systematically apply the iterative discovery process to all root nodes, generating a diverse set of concept-specific circuits that collectively represent the model’s response to the input query. To enhance computational efficiency and prevent duplicated circuit evaluation, we reuse previously computed source node connections and employ recursive techniques to minimize redundant calculations.

\begin{figure*}[!t]
  \centering
     \includegraphics[width=\textwidth]{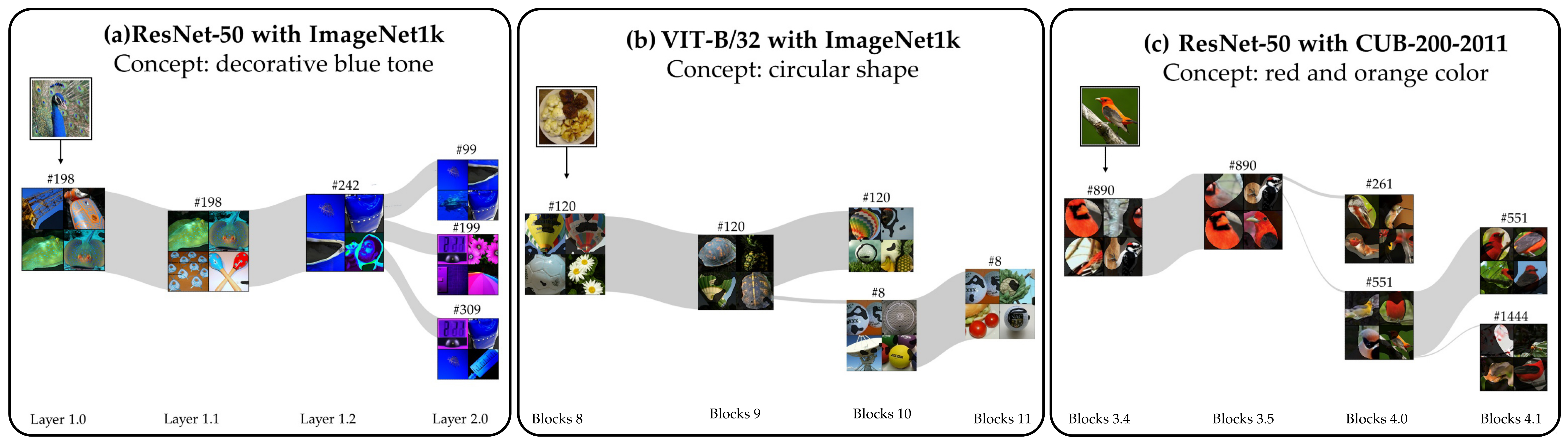}
     \caption{GCC exemplars on various datasets and models. Each node is visualized with its index and 4 highly activated cropped images.}
     \label{fig:qualitative} 
\end{figure*}

\section{Experiments}
\label{sec:exp}

\paragraph{Settings.} This section evaluates whether the proposed method can effectively capture GCC, where each circuit corresponds to a distinct query-related concept. We select five base models—VGG19, ResNet50, ResNet101, MobileNetV3, and ViT—all pretrained on ImageNet1K. Detailed descriptions of the hyperparameter settings are provided in Appendix~\ref{appx:setting}.

\paragraph{Visualization Method.} To illustrate the discovered circuits, we use Sankey diagrams~\cite{schmidt2008sankey}, where link thickness represents the connection strength, computed as described in \Cref{subsec:association}. Each GCC represents a concept-specific circuit for a given query image, with individual nodes denoting neurons. The concept associated with a neuron is inferred from its highly activated samples, with four representative images per node selected from the top 10 highly activated samples in the validation set. To enhance interpretability, we apply threshold-based masking to supporess weak activations, generates a binary mask, and crop high-activation regions. Low activation regions appear in a darker shade to improve visual clarity. See Appendix~\ref{appx:vis_setting} for details.

\subsection{Qualitative Results}

\noindent\textbf{Exploring diverse concepts encoded in the model.}
By identifying GCCs for each root node and merging them, we construct a comprehensive unified circuit representing all concepts related to the query. Figure~\ref{fig:multi-concepts} presents exemplary results for a scoreboard query image, visualizing three of the 17 extracted GCCs, each corresponding to concepts such as the sky background, flags, and clock. To aid interpretation, we also provide manual textual descriptions. Red solid outlines highlight the highly activated regions in the query image with respect to each circuit’s root node. These results highlight our method's ability to disentangle and represent multiple query-related concepts. Notably, this is the first study to construct circuits that capture fine-grained and diverse concepts within a query.

\noindent\textbf{Applying GCC to Different Models and Datasets.}
The proposed algorithm effectively captures hierarchical concept flows across diverse models and datasets, as illustrated in Figure~\ref{fig:qualitative}. In ResNet50, the identified circuit for a peacock image shows a decorative blue tone concept. The initial layer (Layer 1.0) captures broader blue tones, which then refine in subsequent layers (Layer 1.1 and 1.2) to more specific patterns like blue scales and textured blue surfaces, and by Layer 2.0, the concept further specializes into distinctive blue patterns found on decorative objects. For a cauliflower image, ViT-B/32 instead identifies a circuit rooted in a circular shape, transitioning from patterns in a ball, flower, and hot air balloon to tructured forms, such as a turtle shell, and circular objects like cherry tomatoes. Despite architectural differences, both models exhibit a consistent hierarchical progression of concepts, demonstrating GCC’s robustness in capturing meaningful conceptual transformations.

Additionally, the algorithm generalizes well to the CUB-200-2011 dataset~\cite{wah2011caltech}, which includes 200 bird species. The identified circuit traces the evolution of a ``red and orange color" feature, initially linked to a part of the bird but gradually expanding to encompass the entire bird. This result highlights GCC’s ability to capture fine-grained concept evolution while preserving consistent concept flow across datasets. These findings reinforce GCC’s potential as a versatile tool for concept circuit discovery, improving model interpretability and demonstrating its applicability across diverse architectures and datasets.

\begin{figure*}[!t]
  \centering
     \includegraphics[width=\textwidth]{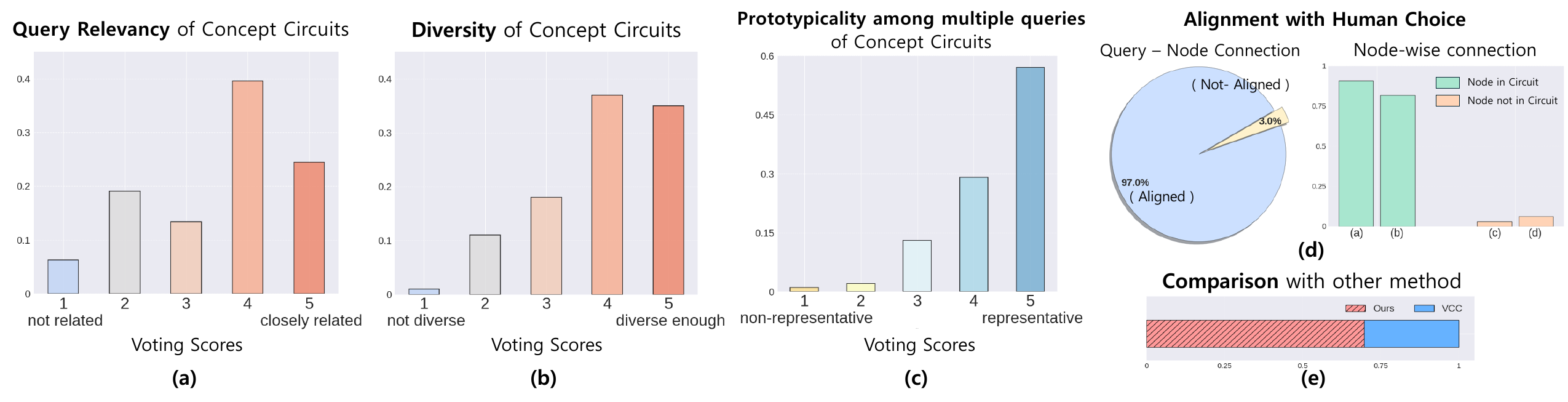}
     \caption{User study results of evaluating GCCs on various aspects:  
\textbf{(a)} Whether the GCC is relevant to the given query. \textbf{(b)} Whether each GCC represents the diverse concepts present in the query. \textbf{(c)} Whether the GCC represents the given multiple queries. \textbf{(d)} Whether users can easily agree with the captured inter-node and query-node connections. \textbf{(e)} Whether our method represents diverse concepts more effectively compared to VCC~\cite{kowal2024visual}.}
     \label{fig:user-study} 
\end{figure*}

\subsection{Quantitative Results}
In this section, we quantitatively evaluate whether the proposed Granular Concept Circuits effectively find concepts in the query. Previous circuit discovery approaches, particularly in the language domain, evaluate circuits based on the key criteria: faithfulness and completeness~\cite{yu2024functional,wang2022interpretability}. Faithfulness refers to a circuit’s ability to perform the task independently, while completeness ensures that all nodes in the circuit are necessary for the task. They assess the faithfulness and completeness of circuits by measuring output corruption after masking the circuit and its complement, respectively. A truly faithful and complete circuit should cause significant corruption when masked, whereas masking its complement should have minimal impact.

Building on this approach, we first evaluate the effectiveness of granular concept circuits by ablating all neurons within each identified circuit and measuring the resulting change in logit outputs. A substantial drop indicates that the circuits faithfully encode query-relevant information. To assess completeness, we ablate the same number of neurons from the complement set (i.e., outside the circuits) and compare the effects. As a baseline, we also include randomly pruned models, ensuring that all pruning conditions involve the same number of ablated neurons for fair comparison. We report average logits across 100 randomly selected ImageNet1k queries. As shown in Table~\ref{tab:quan}, pruning neurons within granular concept circuits leads to a substantial logit drop (8.67$\%_p$ in ResNets), whereas pruning outside the circuits results in a smaller decrease (1.74$\%_p$), even less than random pruning (2.35$\%_p$). These results confirm that our method effectively identifies circuits that are both complete and faithful. The corresponding experimental results for transformer-based models are provided in Appendix~\ref{appx:vit_quan}.

\begin{figure}
    \centering
    \includegraphics[width=\columnwidth]{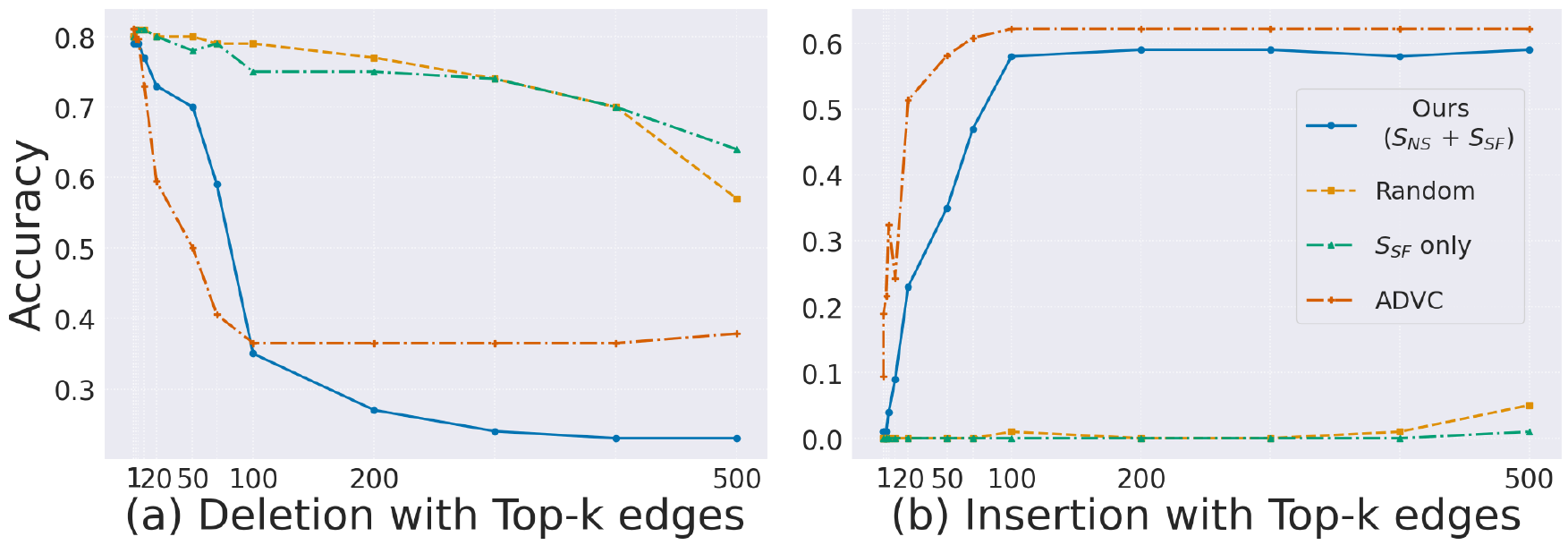 }
    \caption{Deletion and Insertion Curves based on $S_{NS}$ and $S_{SF}$}
    \label{fig:insert-delete}
\end{figure}

To further evaluate the fidelity of edges captured by GCC, we conduct rank-based edge ablation and insertion experiments, following the protocol of \citet{petsiuk2018rise}. Edges are removed or inserted in descending order of their $S_{NS}$ scores. All experiments are performed on the last block of ResNet-50. As in Figure~\ref{fig:insert-delete}, removing top-ranked edges consistently degrades performance, whereas adding them yields clear gains. Although AVCD~\cite{rajaram2024automatic}—a gradient-based, single-circuit discovery method—shows slightly steeper curves, our method constructs fine-grained circuits for individual concepts rather than a single unified circuit tied to class labels. This enables it to extract richer information from the model, resulting in the larger performance drop observed in Fig.~\ref{fig:insert-delete}-(a). 

\begin{table}[!ht]
\centering
\begin{tabular}{c|c|c|c|c|c}
\toprule
                        & R50 & R101 & V19 & M3 & Avg \\ \midrule
Original                & 17.17    & 17.46     & 20.94 & 17.34 & -      \\ \midrule
Random                  & 15.66    & 13.80     & 19.03 & 15.01 & (\mydowntriangle 2.35)      \\ \midrule
\textbf{Ours}                    & \textbf{6.41}     & \textbf{6.18}      & \textbf{12.93} & \textbf{12.95} & (\textbf{\mydowntriangle 8.60})      \\ \midrule \midrule
$\text{Ours}^C$ & 16.12    & 14.58     & 19.93 & 15.88&  (\mydowntriangle 1.74)      \\
\bottomrule
\end{tabular}
\caption{Comparison on average logit drop after ablating neurons with random selection (\textbf{Random}), found circuits (\textbf{Ours}), and non-relevant neurons (\textbf{Ours\textsuperscript{C}).} We abbreviate ResNet50, ResNet101, VGG19 and MobileNetV3.
}
\label{tab:quan}
\end{table}

\subsection{User Study}
In addition to the quantitative results, we conducted a user study with 33 participants, evaluating whether the circuits captured by our method are well-structured and perceptually meaningful to humans. We evaluated the Granular Concept Circuit across three criteria: query relativeness, diversity, and prototypicality. As shown in Figure~\ref{fig:user-study} (a)-(c), the average scores for each criterion were 3.65, 4.0, and 4.45 out of 5, respectively. While all scores indicate that the GCC meets the three criteria, the query relativeness score was slightly lower. This may be because our method captures not only the main part of the query but also background elements and other attributes, resulting in a broader representation beyond the direct focus. In Figure~\ref{fig:user-study}-(d), we assessed the appropriateness of the node-to-node and query-node connections generated by our method by comparing them to human judgment. Furthermore, we compared the GCC with VCC~\cite{kowal2024visual}, and 70$\%$ of users agreed that our method captures more meaningful and diverse concepts than VCC. More than 90$\%$ of users found these connections to be appropriate. The detailed results and experimental settings can be found in Appendix~\ref{appx:user}.

\begin{figure}[!t]
  \centering
     \includegraphics[width=\columnwidth]{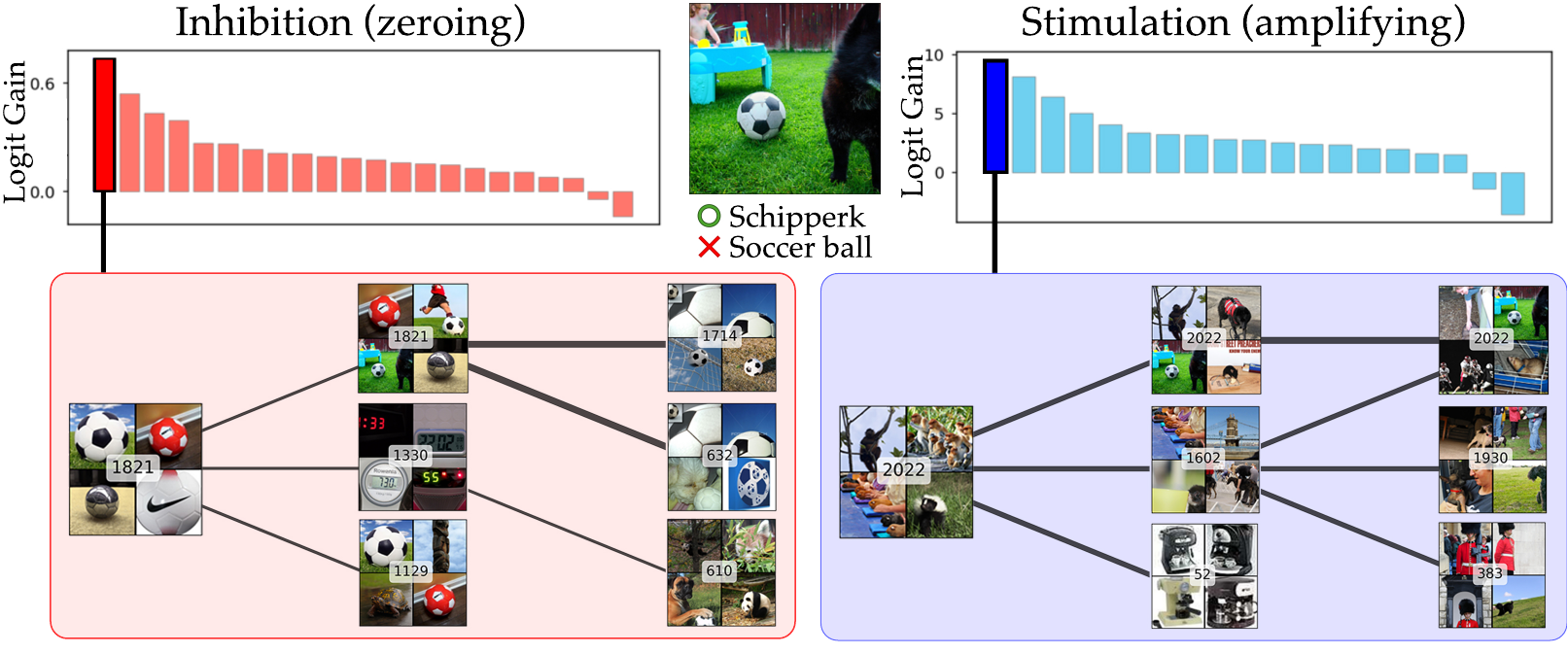}
 \caption{Auditing Misclassification with GCC. The top bar plots show logit gain when GCC constructed on the last residual block of ResNet50, derived from a misclassified "Schipperke" as "Soccer ball" example, are either inhibited (zeroed out) or stimulated (amplified). The bottom panel visualizes the corresponding most influential GCC, highlighting concepts related to both the correct (\textit{soccer ball}) and misclassified (\textit{black objects}) concepts.}
     \label{fig:uscase_misclass} 
\end{figure}

\subsection{Use Cases}
\paragraph{Audit of Misclassified Query.} In instances where the model misclassifies a query, the Granular Concept Circuit can be employed to identify and audit the error. As illustrated in \Cref{fig:uscase_misclass}, the example shows that although the true class of the query image is \textit{schipperke}, the model erroneously predicts it as a \textit{soccer ball}. We first extract the GCC associated with the query from the last residual block of ResNet50. Then, we inhibit (zero out) and amplify (by two times) each neuron within the GCC to analyze its influence on the logit for the true class. As seen in the \textit{Bottom left} panel of Figure \ref{fig:uscase_misclass}, the concepts displayed by the most influential GCC are visually consistent with those typically found in images of the \textit{soccer ball} class. This visual evidence supports the high logit gain observed when inhibiting this \textit{soccer ball}-related GCC, indicating their strong influence on the misclassification. In contrast, the GCC located in the \textit{Bottom right} panel, which encompass \textit{schipperke}-related concepts, show the largest logit gain when stimulated, thereby contributing to the misclassification audit. Overall, this auditing procedure—by first extracting the query-related GCC and then analyzing the logit gain to the true class—facilitates a clear understanding of how the erroneous prediction is associated with specific conceptual representations. Consequently, this approach enhances the reliability of the input and aids in reducing errors in the model's decision-making process.

\begin{figure}[t]
  \centering
     \includegraphics[width=\columnwidth]{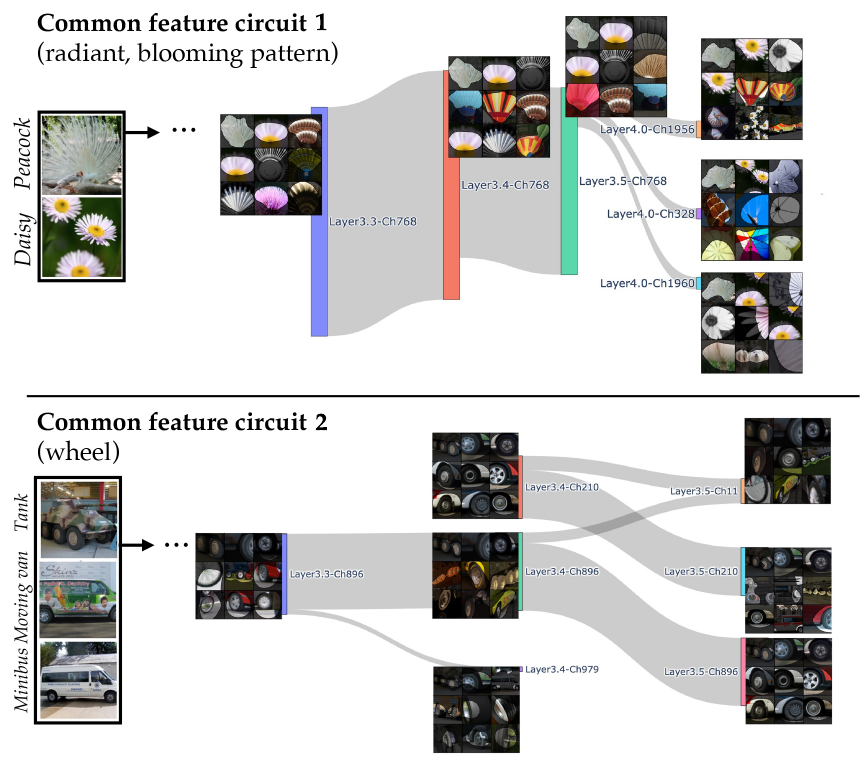}
     \caption{GCC in ResNet50, each capturing shared concepts across multiple queries from different classes.
     }
     \label{fig:usecase_multiple_different_class} 
\end{figure}

\noindent\textbf{Finding Common Concepts Among Multiple Queries.}
Identifying concepts shared across queries from different classes is a challenging task for existing circuit-discovery approaches, particularly methods relying on gradient-based attribution or initial class logits.  In contrast, our approach discovers granular concept circuits by propagating forward from conceptually meaningful root neurons, enabling the effective identification of shared patterns and relationships. As demonstrated in \Cref{fig:usecase_multiple_different_class}, our method identifies a shared “radiant” pattern across fundamentally distinct classes (e.g., daisy and peacock) and similarly captures a common “wheel” concept among various vehicles (tanks, minibuses, moving vans). These results highlight the unique capability of out framework to detect abstract, class-transcending features, providing valuable insights for model interpretability and debugging that may not be achievable by conventional backward circuit discovery approaches.

\section{Conclusion}
\label{sec:discussion}

In this paper, we introduce Granular Concept Circuit (GCC), a novel and effective method for discovering visual circuits that represent specific concepts related to a given query. In spite of its novelty, our approach has certain limitations. As connectivity is evaluated in a model-centric manner, some strongly associated connections may remain difficult to interpret. Moreover, due to the complexity of deep neural networks, a single concept may be distributed across multiple circuit pathways, especially under strict connectivity thresholds. Despite these challenges, we believe our work—being the first to enable fine-grained visual concept circuit discovery—makes a meaningful contribution to AI interpretability research.
\section*{Acknowledgments}
This work was partly supported by the Institute for Information \& Communications Technology Planning \& Evaluation (IITP) grant funded by the Korea government (MSIT) (RS-2019-II190075, Artificial Intelligence Graduate School Support Program (KAIST); RS-2024-00457882, AI Research Hub Project; RS-2022-II220984, Development of Artificial Intelligence Technology for Personalized Plug-and-Play Explanation and Verification of Explanation), and by Center for Applied Research in Artificial Intelligence (CARAI) grant funded by DAPA and ADD (UD230017TD).

{
    \small
    \bibliographystyle{ieeenat_fullname}
    \bibliography{main}

\begin{thebibliography}{42}
\providecommand{\natexlab}[1]{#1}
\providecommand{\url}[1]{\texttt{#1}}
\expandafter\ifx\csname urlstyle\endcsname\relax
  \providecommand{\doi}[1]{doi: #1}\else
  \providecommand{\doi}{doi: \begingroup \urlstyle{rm}\Url}\fi

\bibitem[Achtibat et~al.(2023)Achtibat, Dreyer, Eisenbraun, Bosse, Wiegand, Samek, and Lapuschkin]{achtibat2023attribution}
Reduan Achtibat, Maximilian Dreyer, Ilona Eisenbraun, Sebastian Bosse, Thomas Wiegand, Wojciech Samek, and Sebastian Lapuschkin.
\newblock From attribution maps to human-understandable explanations through concept relevance propagation.
\newblock \emph{Nature Machine Intelligence}, 2023.

\bibitem[Authors(2023)]{anonauthors2024cub200}
Anonymous Authors.
\newblock Cub-200-resnet50 pretrained model, 2023.

\bibitem[Bau et~al.(2017)Bau, Zhou, Khosla, Oliva, and Torralba]{bau2017network}
David Bau, Bolei Zhou, Aditya Khosla, Aude Oliva, and Antonio Torralba.
\newblock Network dissection: Quantifying interpretability of deep visual representations.
\newblock In \emph{Proceedings of the IEEE conference on computer vision and pattern recognition}, 2017.

\bibitem[Bau et~al.(2020)Bau, Zhu, Strobelt, Lapedriza, Zhou, and Torralba]{bau2020units}
David Bau, Jun-Yan Zhu, Hendrik Strobelt, Agata Lapedriza, Bolei Zhou, and Antonio Torralba.
\newblock Understanding the role of individual units in a deep neural network.
\newblock \emph{Proceedings of the National Academy of Sciences}, 2020.

\bibitem[Bennett et~al.(2018)Bennett, Kirby, and Finnerty]{bennett2018rewiring}
Sophie~H Bennett, Alastair~J Kirby, and Gerald~T Finnerty.
\newblock Rewiring the connectome: evidence and effects.
\newblock \emph{Neuroscience \& Biobehavioral Reviews}, 2018.

\bibitem[Chang et~al.(2024)Chang, Kwon, and Choi]{chang2024understanding}
Wonjoon Chang, Dahee Kwon, and Jaesik Choi.
\newblock Understanding distributed representations of concepts in deep neural networks without supervision.
\newblock In \emph{Proceedings of the AAAI Conference on Artificial Intelligence}, 2024.

\bibitem[Conmy et~al.(2023)Conmy, Mavor-Parker, Lynch, Heimersheim, and Garriga-Alonso]{conmy2023towards}
Arthur Conmy, Augustine Mavor-Parker, Aengus Lynch, Stefan Heimersheim, and Adri{\`a} Garriga-Alonso.
\newblock Towards automated circuit discovery for mechanistic interpretability.
\newblock \emph{Advances in Neural Information Processing Systems}, 2023.

\bibitem[Dai et~al.(2021)Dai, Dong, Hao, Sui, Chang, and Wei]{dai2021knowledge}
Damai Dai, Li Dong, Yaru Hao, Zhifang Sui, Baobao Chang, and Furu Wei.
\newblock Knowledge neurons in pretrained transformers.
\newblock \emph{arXiv preprint arXiv:2104.08696}, 2021.

\bibitem[Davison and Smith(1990)]{davison1990models}
Anthony~C Davison and Richard~L Smith.
\newblock Models for exceedances over high thresholds.
\newblock \emph{Journal of the Royal Statistical Society Series B: Statistical Methodology}, 1990.

\bibitem[Dorszewski et~al.(2025)Dorszewski, T{\v{e}}tkov{\'a}, Jenssen, Hansen, and Wickstr{\o}m]{dorszewski2025colors}
Teresa Dorszewski, Lenka T{\v{e}}tkov{\'a}, Robert Jenssen, Lars~Kai Hansen, and Kristoffer~Knutsen Wickstr{\o}m.
\newblock From colors to classes: Emergence of concepts in vision transformers.
\newblock \emph{arXiv preprint arXiv:2503.24071}, 2025.

\bibitem[Fel et~al.(2023)Fel, Picard, Bethune, Boissin, Vigouroux, Colin, Cad{\`e}ne, and Serre]{fel2023craft}
Thomas Fel, Agustin Picard, Louis Bethune, Thibaut Boissin, David Vigouroux, Julien Colin, R{\'e}mi Cad{\`e}ne, and Thomas Serre.
\newblock Craft: Concept recursive activation factorization for explainability.
\newblock In \emph{Proceedings of the IEEE/CVF Conference on Computer Vision and Pattern Recognition}, 2023.

\bibitem[Geva et~al.(2020)Geva, Schuster, Berant, and Levy]{geva2020transformer}
Mor Geva, Roei Schuster, Jonathan Berant, and Omer Levy.
\newblock Transformer feed-forward layers are key-value memories.
\newblock \emph{arXiv preprint arXiv:2012.14913}, 2020.

\bibitem[Ghorbani et~al.(2019)Ghorbani, Wexler, Zou, and Kim]{ghorbani2019towards}
Amirata Ghorbani, James Wexler, James~Y Zou, and Been Kim.
\newblock Towards automatic concept-based explanations.
\newblock \emph{Advances in neural information processing systems}, 2019.

\bibitem[Haan and Ferreira(2006)]{haan2006extreme}
Laurens Haan and Ana Ferreira.
\newblock \emph{Extreme value theory: an introduction}.
\newblock Springer, 2006.

\bibitem[Hebb(2005)]{hebb2005organization}
Donald~Olding Hebb.
\newblock \emph{The organization of behavior: A neuropsychological theory}.
\newblock Psychology press, 2005.

\bibitem[Hernandez et~al.(2021)Hernandez, Schwettmann, Bau, Bagashvili, Torralba, and Andreas]{hernandez2021natural}
Evan Hernandez, Sarah Schwettmann, David Bau, Teona Bagashvili, Antonio Torralba, and Jacob Andreas.
\newblock Natural language descriptions of deep visual features.
\newblock In \emph{International Conference on Learning Representations}, 2021.

\bibitem[Hinton(1984)]{hinton1984distributed}
Geoffrey~E Hinton.
\newblock Distributed representations.
\newblock 1984.

\bibitem[Hinton(1986)]{hinton1986learning}
Geoffrey~E Hinton.
\newblock Learning distributed representations of concepts.
\newblock In \emph{Proceedings of the Annual Meeting of the Cognitive Science Society}, 1986.

\bibitem[Jeong et~al.(2021)Jeong, Cho, Kim, Oh, Kang, Yoo, Lee, and Han]{jeong2021synaptic}
Yire Jeong, Hye-Yeon Cho, Mujun Kim, Jung-Pyo Oh, Min~Soo Kang, Miran Yoo, Han-Sol Lee, and Jin-Hee Han.
\newblock Synaptic plasticity-dependent competition rule influences memory formation.
\newblock \emph{Nature communications}, 2021.

\bibitem[Kalibhat et~al.(2023)Kalibhat, Bhardwaj, Bruss, Firooz, Sanjabi, and Feizi]{kalibhat2023identifying}
Neha Kalibhat, Shweta Bhardwaj, C~Bayan Bruss, Hamed Firooz, Maziar Sanjabi, and Soheil Feizi.
\newblock Identifying interpretable subspaces in image representations.
\newblock In \emph{International Conference on Machine Learning}. PMLR, 2023.

\bibitem[Kim et~al.(2018)Kim, Wattenberg, Gilmer, Cai, Wexler, Viegas, et~al.]{kim2018interpretability}
Been Kim, Martin Wattenberg, Justin Gilmer, Carrie Cai, James Wexler, Fernanda Viegas, et~al.
\newblock Interpretability beyond feature attribution: Quantitative testing with concept activation vectors (tcav).
\newblock In \emph{International conference on machine learning}, pages 2668--2677. PMLR, 2018.

\bibitem[Kowal et~al.(2024)Kowal, Wildes, and Derpanis]{kowal2024visual}
Matthew Kowal, Richard~P Wildes, and Konstantinos~G Derpanis.
\newblock Visual concept connectome (vcc): Open world concept discovery and their interlayer connections in deep models.
\newblock In \emph{Proceedings of the IEEE/CVF Conference on Computer Vision and Pattern Recognition}, pages 10895--10905, 2024.

\bibitem[Marks et~al.(2024)Marks, Rager, Michaud, Belinkov, Bau, and Mueller]{marks2024sparse}
Samuel Marks, Can Rager, Eric~J Michaud, Yonatan Belinkov, David Bau, and Aaron Mueller.
\newblock Sparse feature circuits: Discovering and editing interpretable causal graphs in language models.
\newblock \emph{arXiv preprint arXiv:2403.19647}, 2024.

\bibitem[Meng et~al.(2022)Meng, Bau, Andonian, and Belinkov]{meng2022locating}
Kevin Meng, David Bau, Alex Andonian, and Yonatan Belinkov.
\newblock Locating and editing factual associations in gpt.
\newblock \emph{Advances in neural information processing systems}, 2022.

\bibitem[Oikarinen and Weng(2023)]{oikarinen2022clip}
Tuomas~P. Oikarinen and Tsui{-}Wei Weng.
\newblock Clip-dissect: Automatic description of neuron representations in deep vision networks.
\newblock In \emph{The Eleventh International Conference on Learning Representations, {ICLR} 2023, Kigali, Rwanda, May 1-5, 2023}. OpenReview.net, 2023.

\bibitem[Olah et~al.(2020)Olah, Cammarata, Schubert, Goh, Petrov, and Carter]{olah2020overview}
Chris Olah, Nick Cammarata, Ludwig Schubert, Gabriel Goh, Michael Petrov, and Shan Carter.
\newblock An overview of early vision in inceptionv1.
\newblock \emph{Distill}, 2020.

\bibitem[Petsiuk et~al.(2018)Petsiuk, Das, and Saenko]{petsiuk2018rise}
Vitali Petsiuk, Abir Das, and Kate Saenko.
\newblock Rise: Randomized input sampling for explanation of black-box models.
\newblock \emph{arXiv preprint arXiv:1806.07421}, 2018.

\bibitem[Pickands~III(1975)]{pickands1975statistical}
James Pickands~III.
\newblock Statistical inference using extreme order statistics.
\newblock \emph{the Annals of Statistics}, 1975.

\bibitem[Rajaram et~al.(2024)Rajaram, Chowdhury, Torralba, Andreas, and Schwettmann]{rajaram2024automatic}
Achyuta Rajaram, Neil Chowdhury, Antonio Torralba, Jacob Andreas, and Sarah Schwettmann.
\newblock Automatic discovery of visual circuits.
\newblock \emph{arXiv preprint arXiv:2404.14349}, 2024.

\bibitem[Rissman and Wagner(2012)]{rissman2012distributed}
Jesse Rissman and Anthony~D Wagner.
\newblock Distributed representations in memory: insights from functional brain imaging.
\newblock \emph{Annual review of psychology}, 2012.

\bibitem[Rumelhart et~al.(1986)Rumelhart, McClelland, Group, et~al.]{rumelhart1986parallel}
David~E Rumelhart, James~L McClelland, PDP~Research Group, et~al.
\newblock \emph{Parallel distributed processing, volume 1: Explorations in the microstructure of cognition: Foundations}.
\newblock The MIT press, 1986.

\bibitem[Schmidt and Polleux(2022)]{schmidt2022genetic}
Ewoud~RE Schmidt and Franck Polleux.
\newblock Genetic mechanisms underlying the evolution of connectivity in the human cortex.
\newblock \emph{Frontiers in Neural Circuits}, 2022.

\bibitem[Schmidt(2008)]{schmidt2008sankey}
Mario Schmidt.
\newblock The sankey diagram in energy and material flow management: part ii: methodology and current applications.
\newblock \emph{Journal of industrial ecology}, 2008.

\bibitem[Tukey et~al.(1977)]{tukey1977exploratory}
John~Wilder Tukey et~al.
\newblock \emph{Exploratory data analysis}.
\newblock Springer, 1977.

\bibitem[Wah et~al.(2011)Wah, Branson, Welinder, Perona, and Belongie]{wah2011caltech}
Catherine Wah, Steve Branson, Peter Welinder, Pietro Perona, and Serge Belongie.
\newblock The caltech-ucsd birds-200-2011 dataset.
\newblock 2011.

\bibitem[Wang et~al.(2022{\natexlab{a}})Wang, Lee, and Qi]{wang2022hint}
Andong Wang, Wei-Ning Lee, and Xiaojuan Qi.
\newblock Hint: Hierarchical neuron concept explainer.
\newblock In \emph{Proceedings of the IEEE/CVF Conference on Computer Vision and Pattern Recognition}, 2022{\natexlab{a}}.

\bibitem[Wang et~al.(2022{\natexlab{b}})Wang, Variengien, Conmy, Shlegeris, and Steinhardt]{wang2022interpretability}
Kevin Wang, Alexandre Variengien, Arthur Conmy, Buck Shlegeris, and Jacob Steinhardt.
\newblock Interpretability in the wild: a circuit for indirect object identification in gpt-2 small.
\newblock \emph{arXiv preprint arXiv:2211.00593}, 2022{\natexlab{b}}.

\bibitem[Wang et~al.(2025)Wang, Liu, Shi, Li, Pang, Yang, Yu, and Ren]{wang2025discovering}
Yifan Wang, Yifei Liu, Yingdong Shi, Changming Li, Anqi Pang, Sibei Yang, Jingyi Yu, and Kan Ren.
\newblock Discovering influential neuron path in vision transformers.
\newblock \emph{arXiv preprint arXiv:2503.09046}, 2025.

\bibitem[Wood(1996)]{wood1996neuroscience}
Isaac~K Wood.
\newblock Neuroscience: Exploring the brain: By mark f. bear, barry w. connors and michael a. paradiso. baltimore: Williams \& wilkins, 1996. pp. 666, 1996.

\bibitem[Xin et~al.(2019)Xin, Lin, and Yu]{xin2019part}
Ji Xin, Jimmy Lin, and Yaoliang Yu.
\newblock What part of the neural network does this? understanding lstms by measuring and dissecting neurons.
\newblock In \emph{Proceedings of the 2019 Conference on Empirical Methods in Natural Language Processing and the 9th International Joint Conference on Natural Language Processing (EMNLP-IJCNLP)}, pages 5823--5830, 2019.

\bibitem[Yu et~al.(2024)Yu, Niu, Zhu, and Penn]{yu2024functional}
Lei Yu, Jingcheng Niu, Zining Zhu, and Gerald Penn.
\newblock Functional faithfulness in the wild: Circuit discovery with differentiable computation graph pruning.
\newblock \emph{arXiv preprint arXiv:2407.03779}, 2024.

\bibitem[Zhou et~al.(2018)Zhou, Sun, Bau, and Torralba]{zhou2018revisiting}
Bolei Zhou, Yiyou Sun, David Bau, and Antonio Torralba.
\newblock Revisiting the importance of individual units in cnns via ablation.
\newblock \emph{arXiv preprint arXiv:1806.02891}, 2018.

\end{thebibliography}
}

\appendix
\clearpage
\renewcommand{\thefigure}{\Alph{figure}}
\renewcommand{\thetable}{\Alph{table}}
\renewcommand{\thesection}{\Alph{section}}
\setcounter{page}{1}
\maketitlesupplementary

\section{Experimental Settings}
\label{appx:setting}

\subsection{Model Configuration}
Our experiments cover four CNNs (ResNet‑50/101, VGG‑19, MobileNetV3) and three transformer-based models (ViT‑B/32, Swin Transformer, CLIP‑ViT). For brevity, we refer to ResNet‑50, ResNet‑101, VGG‑19, and MobileNetV3 as R50, R101, V19, and M3, respectively. \Cref{tab:layers} details the network layers analyzed for identifying concept circuits. We utilize model architectures and their corresponding pretrained weights from Torchvision on ImageNet, except for the CUB-200-2011 dataset, where we employ the anonauthors/cub200-resnet50 pretrained weights available on the Hugging Face Hub~\cite{anonauthors2024cub200}. In transformer-based models, the feed-forward network (FFN) is known to serve as a memory mechanism, owing to its structural resemblance to the attention mechanism. Moreover, previous studies~\cite{geva2020transformer,wang2025discovering} indicate that it captures a range of high-level, human-interpretable concepts. Based on this, we analyze the activations from the first linear layer of the FFN in each Transformer block.

\begin{table}[!ht]
\centering
\begin{tabularx}{\columnwidth}{l|X}
\toprule
\textbf{Model} & \textbf{Layer Names} \\
\midrule
R50 & \texttt{layer1.0, \dots, layer1.2, layer2.0, \dots, layer2.3, layer3.0, \dots, layer3.5, layer4.0, \dots, layer4.2} \\
\addlinespace 
R101 & \texttt{layer1.0, \dots, layer1.2, layer2.0, \dots, layer2.3, layer3.0, \dots, layer3.22, layer4.0, \dots, layer4.2} \\
\addlinespace
V19 & \texttt{1, 3, 6, 8, 11, 13, 15, 18, 20, 22, 25, 27, 29, 32, 34, 36} \\
\addlinespace
M3 & \texttt{0.0, 1.0, 1.1, 2.0, 2.1, 2.2, 3.0, 3.1, 3.2, 3.3, 4.0, 4.1, 5.0, 5.1, 5.2, 6.0} \\
\addlinespace
ViT-B & \texttt{encoder.layers.0, \dots, encoder.layers.11} \\
\addlinespace
CLIP-ViT & \texttt{visual.transformer.resblocks.0, \dots, resblocks.11} \\
\addlinespace
Swin-T & \texttt{features.SwinTransformerBlock.0, \dots, SwinTransformerBlock.11} \\
\bottomrule
\end{tabularx}
\caption{Network layers used for constructing Granular Concept Circuits.}
\label{tab:layers}
\end{table}

\subsection{Root Node Selection} 

The choice of root nodes determines the conceptual focus of the Granular Concept Circuits (GCCs). We define root nodes as neurons that rank among the top 1\% in activation across the entire dataset. However, this threshold can be adjusted for broader exploration. For instance, selecting the top 10\% highly activated neurons increases the number of discovered circuits, allowing for a wider variety of conceptual representations. While this approach enhances diversity, it introduces a trade-off: if a root node has lower activation levels in the given query, the retrieved circuit may be less relevant to the query itself. Therefore, the selection of root nodes should be carefully tuned to balance concept diversity and query relevance.

\subsection{Parameter Selections.}
\label{sec:threshold}

\paragraph{Selections for $\tau_{NS}$.} To evaluate connectivity between two nodes, we introduce two scores, one of which is the Neuron Sensitivity Score ($S_{NS}$). This score quantifies the influence of a neuron (source node) in the current layer by measuring changes in the next layer when the source node is zero-masked. If a particular node in the next layer experiences an exceptionally large decrease, we consider it strongly associated with the source node.

To identify such significant changes, we recommend using the Peak Over Threshold (POT) method, a statistical approach from Extreme Value Theory (EVT) commonly used in anomaly detection. POT models the tail behavior of a distribution by focusing on values exceeding a high threshold, without requiring assumptions about the underlying distribution. As shown in Figure~\ref{fig:score}, our score distribution is typically right-skewed, allowing us to identify sufficiently large score values as positive anomalies. Specifically, POT captures extreme events by selecting all observations above a predefined threshold, with exceedances modeled using the Generalized Pareto Distribution (GPD). Choosing an appropriate threshold is crucial—setting it too low includes non-extreme values, while setting it too high limits data availability. We examine the effect of the POT threshold on constructing GCCs by testing thresholds at 95, 90, 80, 70, and 60 percentiles. A looser threshold (e.g., 60) captures more nodes, while stricter thresholds may focus on the most significant ones. We quantitatively compare logit changes when attenuating GCCs with different thresholds, using the same settings as Table~\ref{tab:quan}. 

Contrary to expectations, the degree of logit drop remains largely consistent across different POT thresholds. We hypothesize that this is because only a small portion of the network is actively engaged with a given query, while the rest is redundant. Since looser POT thresholds do not significantly improve performance but may introduce redundant circuits and increase computational cost, we recommend using a POT threshold between 90 and 95. While the optimal parameter may vary depending on the dataset or model, our empirical results suggest that thresholds in this range provide sufficiently strong performance. Note that the threshold for $S_{NS}$ can also be determined using other well-established statistical methods, such as Interquartile Range-based outlier detection~\cite{tukey1977exploratory}.

\begin{table}[!ht]
\centering
\resizebox{\columnwidth}{!}{ 
\begin{tabular}{c|c|c|c|c|c|c}
\toprule
POT   & 95 & 90  & 80 & 70 & 60 & AVG \\ 
\midrule
Original  & 17.17 & 16.77   & 17.28 & 16.71 & 16.52 & -\\ 
\midrule
Random    & 15.66    & 15.15  & 15.61 & 14.81 & 14.54 & (\mydowntriangle 1.74)  \\ 
\midrule
\textbf{Ours} & \textbf{6.41}  & \textbf{6.60} & \textbf{10.42} & \textbf{6.92} & & (\textbf{\mydowntriangle 5.66}) \\ 
\midrule\midrule
$\text{Ours}^C$ & 16.12 & 15.41   & 16.12 & 15.15 & 15.43&  (\mydowntriangle 1.24)   \\
\bottomrule
\end{tabular}
}
\caption{Logit change comparison using threshold values estimated by POT in ResNet50.}
\label{tab:quan_appx}
\end{table}

\paragraph{Selections for $\tau_{SF}$.} The Semantic Flow Score ($S_{SF}$) ensures that information from the source node is meaningfully shared with the connected node. Unlike the Neuron Sensitivity Score ($S_NS$), which directly measures connectivity strength, the Semantic Flow Score helps filter out spuriously aligned nodes among those with high $S_NS$ values.
To achieve this, we introduce a threshold ($\tau_{SF}$) to control the required level of semantic alignment. A stricter $\tau_{SF}$ ensures that only genuinely related nodes are retained, while a looser $\tau_{SF}$ may weaken the constraint. This balance is particularly important in complex models, which often exhibit polysemantic characteristics, where an overly high threshold could inadvertently remove valid connections.
To address this trade-off, we set $\tau_{SF}$ as the average value across all nodes in the target layer, ensuring that only nodes with above-average semantic similarity to the source node are retained. This approach maintains a necessary level of information sharing while allowing the threshold to adapt flexibly based on model and layer characteristics.

\subsection{Visualization strategy} 
\label{appx:vis_setting}
To reveal the specific visual features that contribute to a neuron's activation and enable a precise analysis of learned representations, we apply cropping and masking techniques based on the activation map of a given sample and neuron. The activation map is upsampled to match the input query image size, and Gaussian blur is applied to suppress noise and improve spatial coherence. The processed activation map is then thresholded to generate a binary mask, isolating highly activated regions. The largest connected region is identified using a bounding box, which is subsequently used to crop and standardize the visualization. Finally, less activated areas are darkened by overlaying the binary mask with lower transparency onto the original image, improving interpretability. 

\section{Quantitative Results on Transformers}
\label{appx:vit_quan}

In line with the quantitative results on convolution-based models (Table~\ref{tab:quan}), we evaluate performance degradation in transformer-based models by ablating the discovered circuits. Unlike the previous setting where we measured logit drops, we use accuracy drop as the evaluation metric here, as the transformer architecture includes layer normalization, which can suppress logit differences while still affecting the final decision. The results are presented in Table~\ref{tab:vit_quan}. For CLIP-ViT, classification is based on the text embedding with the highest cosine similarity to the image embedding. The results show that ablating our identified circuits leads to a substantial accuracy drop of $33.85\%_p$, whereas randomly ablating the same number of nodes results in only a $0.81\%_p$ decrease. This indicates that our method can be effectively extended to transformer-based models.

\begin{table}[!ht]
\centering
{\small
\begin{tabular}{c|c|c|c|c}
\toprule
                        & ViT & Swin-T & CLIP-ViT & Avg \\ \midrule
Original                & 76.61    & 81.92     & 62.19 & 73.57    \\ \midrule
Random                  & 77.19    & 81.34     & 59.75 & 72.76 (\mydowntriangle 0.81)   \\ \midrule
\textbf{Ours}           & \textbf{58.47}  & \textbf{36.92}  & \textbf{23.78} & \textbf{39.72} (\mydowntriangle 33.85)  \\ \bottomrule
\end{tabular}}
\caption{Impact of circuit ablation on ViT and its variants.}

\label{tab:vit_quan}
\end{table}

\section{User-Study}
\label{appx:user}

\begin{figure*}[!ht]
  \centering
     \includegraphics[width=0.875\textwidth]{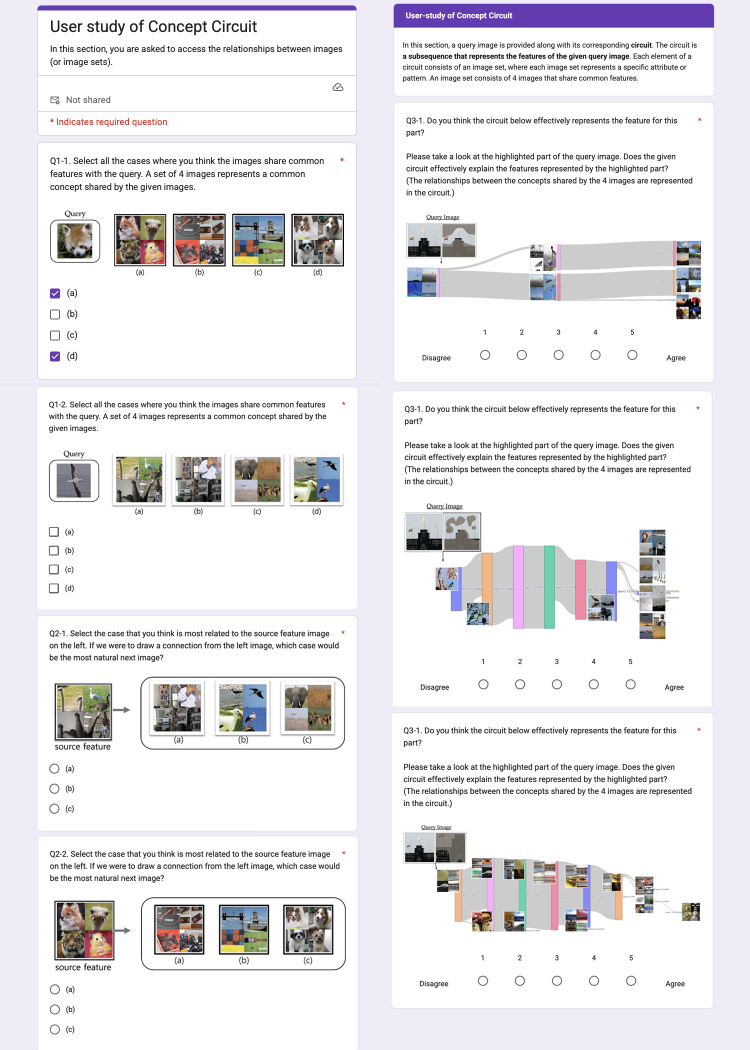}
    \caption{User study questions (Part 1). The questions are presented in a sequential order from top-left to bottom-left, followed by top-right to bottom-right. Participants were instructed to read each question carefully and select their responses based on the given options.}
     \label{fig:user_study_p1}
\end{figure*}

\begin{figure*}[!ht]
  \centering
     \includegraphics[width=0.88\textwidth]{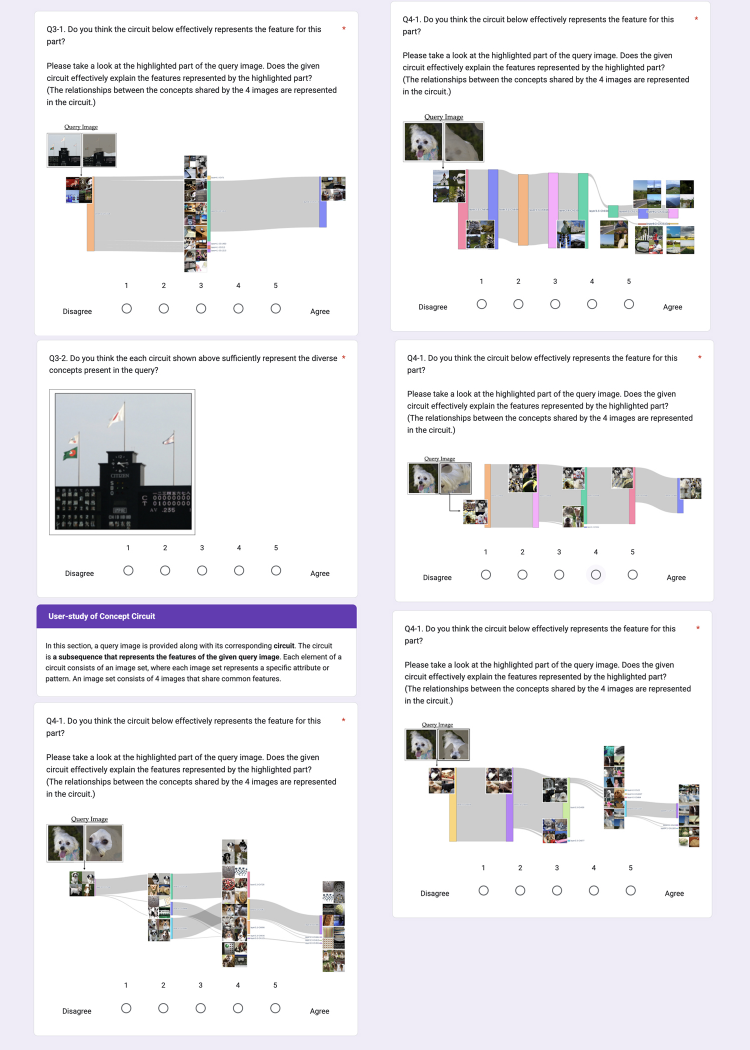}
     \caption{User study questions (Part 2). Continuation of the user study evaluation, following the same structure as \Cref{fig:user_study_p1}.}
     \label{fig:user_study_p2}
\end{figure*}

\begin{figure*}[!ht]
  \centering
     \includegraphics[width=0.9\textwidth]{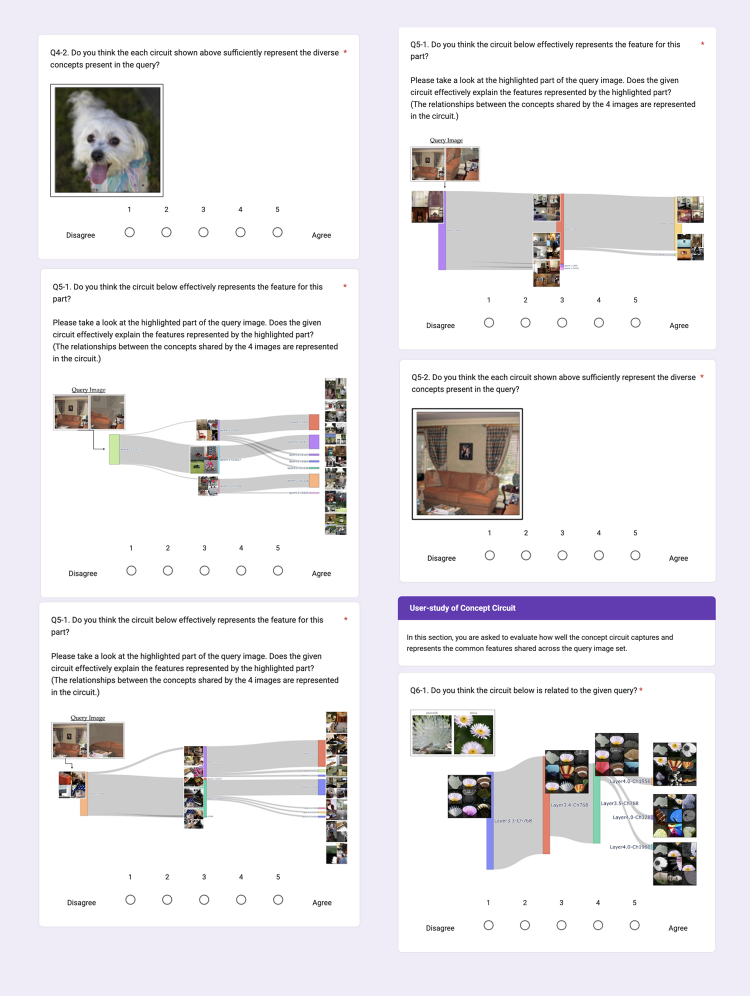}
     \caption{User study questions (Part 3). Continuation of the user study evaluation, following the same structure as \Cref{fig:user_study_p2}.}
     \label{fig:user_study_p3}
\end{figure*}

\begin{figure*}[!ht]
  \centering
     \includegraphics[width=0.95\textwidth]{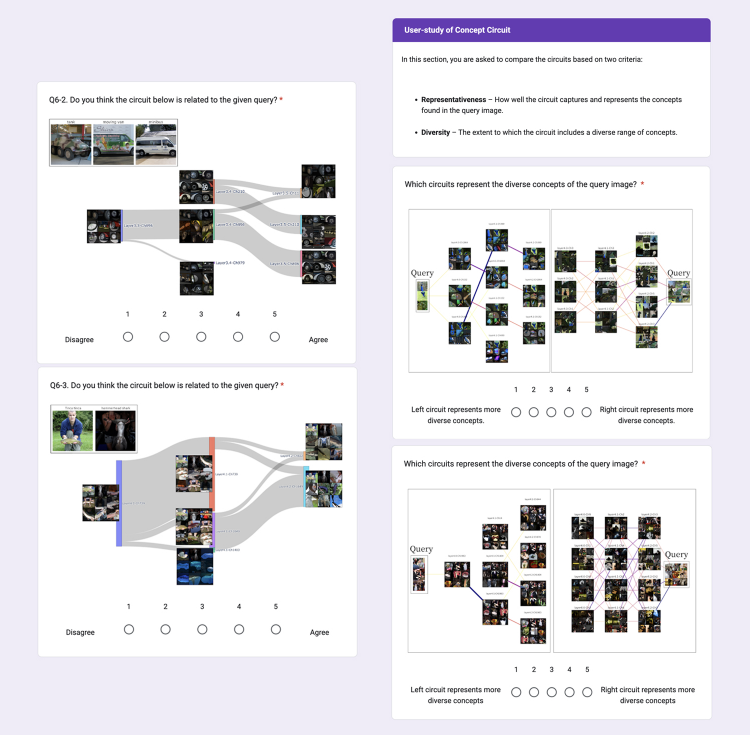}
     \caption{User study questions (Part 4). Final set of evaluation questions in the user study.}
     \label{fig:user_study_p4}
\end{figure*}

To evaluate the interpretability and effectiveness of Granular Concept Circuits (GCC), we conducted a user study with 33 participants. The objective was to assess whether GCC visualizations offer meaningful insights into model behavior and enhance human understanding of circuit representations. \Cref{fig:user_study_p1} to \Cref{fig:user_study_p4} illustrate the questions presented in the user study, while the corresponding responses are summarized in \Cref{fig:user-study}.

\clearpage

 \clearpage

 \clearpage

\clearpage
 
\end{document}